\documentclass[13pt,twocolumn,letterpaper]{article}

\usepackage{cvpr}
\usepackage{times}
\usepackage{epsfig}
\usepackage{graphicx}
\usepackage{amsmath}
\usepackage{amssymb}
\usepackage{ctable}

\usepackage[pagebackref=true,breaklinks=true,letterpaper=true,colorlinks,bookmarks=false]{hyperref}

\cvprfinalcopy %

\def\cvprPaperID{4880} %

\ifcvprfinal\fi
\begin{document}

\title{End-to-end Interpretable Neural Motion Planner}

\author{
  Wenyuan Zeng$^{1,2}$\thanks{denotes equal contribution.} \quad Wenjie Luo$^{1,2*}$\\
  Simon Suo$^{1,2}$ \quad Abbas Sadat$^{1}$ \quad Bin Yang$^{1,2}$ \quad Sergio Casas$^{1,2}$ \quad Raquel Urtasun$^{1,2}$\\
 $^{1}$Uber Advanced Technologies Group \quad $^{2}$University of Toronto\\
 \small\texttt{\{wenyuan,wenjie,suo,byang,sergio,urtasun\}@cs.toronto.edu, abbas@uber.com}
}

\maketitle
\thispagestyle{empty}

\begin{abstract}

In this paper, we propose a neural motion planner (NMP) for learning to drive autonomously in complex urban scenarios that include traffic-light handling, yielding, and interactions with multiple road-users.
Towards this goal, we design a holistic model that takes as input raw LIDAR data and a HD map and produces  interpretable intermediate representations in the form of 3D detections and their future trajectories, as well as a cost volume defining the goodness of each position that the self-driving car  can take within the planning horizon. We then sample a set of diverse physically possible trajectories and choose the one with the minimum learned cost.
Importantly, our cost volume is able to naturally capture  multi-modality.
We demonstrate the effectiveness of our approach in real-world driving data captured  in several cities in North America. Our experiments show that
the learned cost volume can generate safer planning than all the baselines.

\end{abstract}

\section{Introduction}

Self-driving vehicles (SDVs) are going to revolutionize the way we live.
Building reliable  SDVs at scale is, however, not a solved problem.
As is the case in many application domains, the field of autonomous driving has been transformed in the past few years by the success of deep learning.
Existing approaches that leverage this technology can be characterized into two main frameworks:  end-to-end driving and traditional engineering stacks.

\begin{figure*}[t]
\vspace{-0.7cm}
\begin{center}
  \includegraphics[height=8cm]{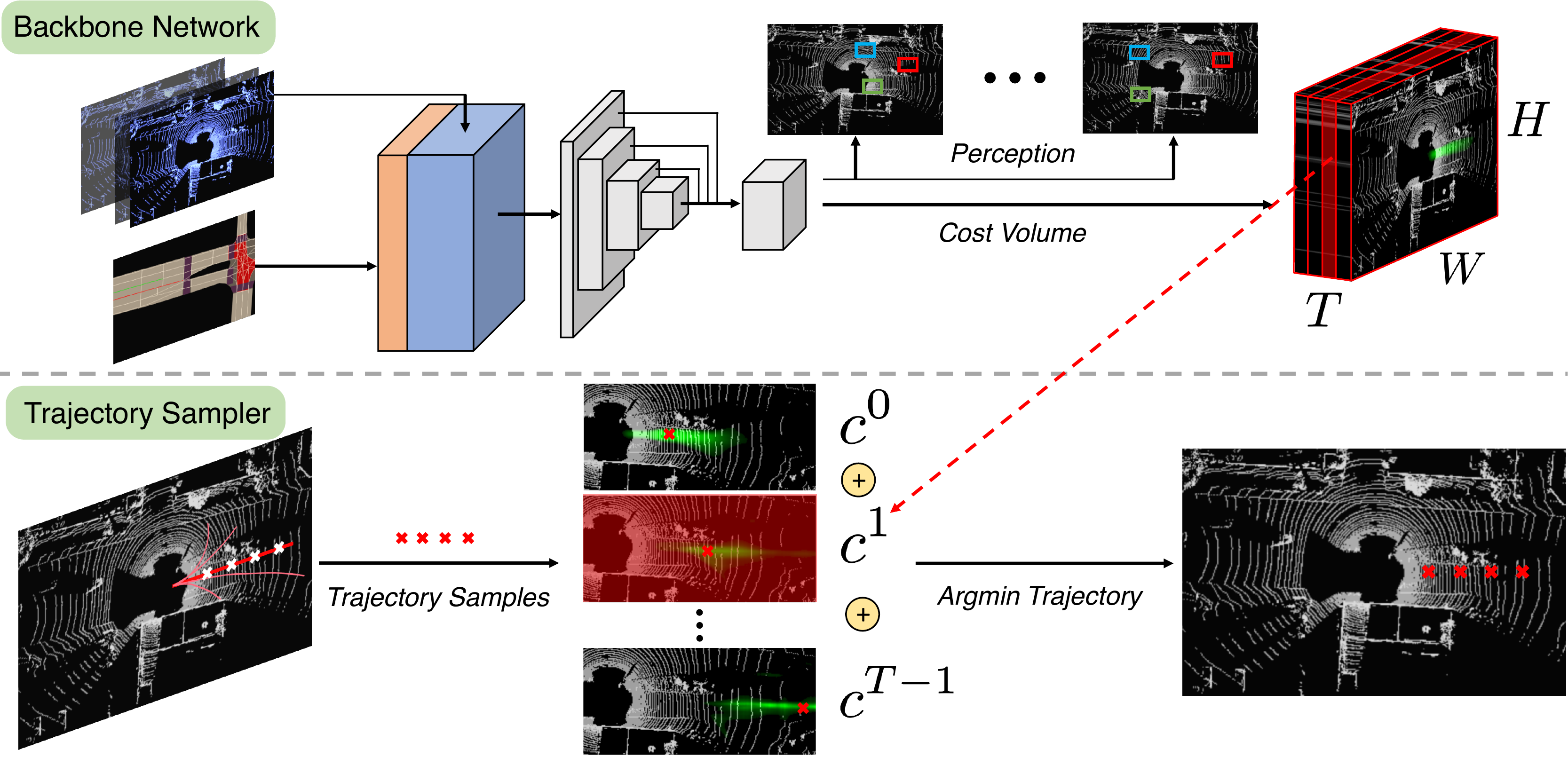}
\end{center}
\vspace{-0.2cm}
\caption{Our end-to-end interpretable neural motion planner (NMP). Backbone network takes LiDAR data and maps as inputs, and outputs bounding boxes of other actors for future timesteps (perception), as well as a cost volume for planning with $T$ filters. Next, for each trajectory proposal from the sampler, its cost is indexed from different filters of the cost volume and summed together. The trajectory with the minimal cost will be our final planning.}
\label{fig:model}
\end{figure*}

 {\it End-to-end driving} approaches \cite{bojarski2016end,pomerleau1989alvinn}  take the output of the sensors (e.g., LiDAR, images) and use it as input to a neural net that outputs  control signals, e.g., steering command and acceleration.
 The main benefit of this framework is its simplicity  as only a few lines of code can build a model and  labeled training data can be
 easily obtained automatically by recording human driving under a SDV platform.
 In practice, this approach suffers from the compounding error due to the nature of self-driving control being a sequential decision problem, and requires massive amounts of data to generalize.
Furthermore, interpretability is difficult to obtain for analyzing the mistakes of the network. It is also hard to incorporate sophisticated prior knowledge about the scene, e.g. that vehicles should not collide.

 In contrast, most self-driving car companies, utilize a {\it traditional engineering stack}, where the problem is divided into subtasks: perception, prediction, motion planning and control. Perception is in charge of estimating all actors' positions and motions, given the current and past evidences. This involves solving tasks such as 3D object detection and tracking. Prediction\footnote{We'll use \textit{prediction} and \textit{motion forecasting} interchangeably.}, on the other hand, tackles the problem of estimating the future positions of all actors as well as their intentions (e.g., changing lanes, parking). Finally, motion planning takes the output from previous stacks and generates a safe trajectory for the SDV to execute via a control system.
This framework has interpretable intermediate representations by construction, and prior knowledge can be easily exploited, for example in the form of high definition maps (HD maps).

However, solving each of these sub-tasks is not only hard, but also may lead to a sub-optimal overall system performance.
Most self-driving companies have large engineering teams working on each sub-problem in isolation, and they train each sub-system with a task specific objective. As a consequence, an advance in one sub-system does not easily translate to an overall system performance improvement. For instance, 3D detection tries to maximize AP, where each actor has the same weight. However, in a driving scenario, high-precision detections of near-range actors who may influence the SDV motion, e.g. through interactions (cutting in, sudden stopping), is more critical.
 In addition, uncertainty estimations are difficult to propagate and computation is not shared among different sub-systems. This leads to longer reaction times of the SDV and make the overall system less reliable.

 In this paper we bridge the gap between these two frameworks.
Towards this goal, we propose the first end-to-end learnable and interpretable motion planner. %
Our model takes as input LiDAR point clouds and a HD map,  and produces  interpretable intermediate representations in the form of 3D detections and their future trajectories. Our final output representation is a space-time cost volume that represents the ``goodness" of each location that the SDV can take within a planning horizon.
Our planner then samples a set of diverse and feasible trajectories, and selects the one with the minimum learned cost for execution.
Importantly, the non-parametric cost volume is able to capture the uncertainty and multi-modality in possible SDV trajectories, e.g changing lane v.s keeping lane.

We demonstrate the effectiveness of our approach in real world driving data captured  in several cities in North America. Our experiments show that our model provides good interpretable representations, and shows better performance. Specifically for detection and motion forecasting, our model outperforms recent neural architectures specifically designed on these tasks. For motion planning, our model generates safer planning compared to the  baselines. %

\section{Related Work}

\textbf{Imitation Learning:} Imitation learning (IL) uses expert demonstrations to directly learn a policy that maps  states to actions.
IL for self-driving vehicles was introduced in the pioneering work of \cite{pomerleau1989alvinn} where a direct mapping from the sensor data to steering angle and acceleration is learned.
\cite{bojarski2016end} follows the similar philosophy.
In contrast, with the help of a high-end driving simulator \cite{dosovitskiy2017carla}, Codevilla \etal \cite{codevilla2018end} exploit conditional models with additional high-level commands such as \textit{continue}, \textit{turn-left}, \textit{turn-right}.
Muller \etal \cite{muller2018driving} incorporate road segmentation as intermediate representations, which are then converted into steering commands. %
In practice, IL approaches suffer from the compounding error due to the nature of self-driving control being a sequential decision problem.
Furthermore, these approaches require massive amount of data, and generalize poorly, e.g., to situations drifting out of lane.

\textbf{RL \& IRL:} Reinforcement learning (RL)  is a natural fit for sequential decision problems as it considers the interactions between the environment and the agent (a self-driving car in this case). Following the  success of Alpha GO \cite{silver2017mastering},  RL has been applied to self-driving in \cite{kendall2018learning,pan2017virtual}.
On the other hand, the inverse reinforcement learning (IRL) looks at learning the reward function for a given task. \cite{wulfmeier2015maximum, ziebart2008maximum} develop IRL algorithms to learn drivable region for self-driving cars. \cite{rhinehart2018r2p2} further infers possible trajectories with a symmetrical cross-entropy loss.
However, all these approaches have only been tested on simulated datasets or small real-world datasets, and it is unclear if RL and IRL can scale to more realistic settings.
Furthermore, these methods do not produce interpretable representations, which are desirable in safety critical applications.

\textbf{Optimization Based Planners:}
Motion planning has long been treated as an independent task  that uses the outputs of perception and prediction modules to formulate an optimization problem, usually by manually engineering a cost function \cite{buehler2009darpa, fan2018baidu, montemerlo2008junior, ziegler2014trajectory}.
The preferred trajectory is then generated by minimizing this cost function.
In practice, to simplify the optimization problem, many approaches assume the objective to be quadratic  \cite{8317745}, decompose lateral and longitudinal planning as two tasks \cite{ajanovic2018search, fan2018baidu} or represent the search space into speed and path \cite{fraichard1993path, kant1986toward}.
In \cite{ajanovic2018search} A* is used to search the space of possible motion.
Similarly, the \textit{Baidu} motion planner \cite{fan2018baidu} uses dynamic programming to find an approximate path and speed profile.
In \cite{ziegler2014trajectory}, the trajectory planning problem is formulated as continuous optimization and used in practice to demonstrate 100km of autonomous driving.
In sampling-based approaches, a set of trajectories is generated and evaluated against a predefined cost, among which, the one with minimum cost is  chosen  \cite{7535557, werling2010optimal}. Such approaches are attractive since they are highly parallelizable \cite{mcnaughton2011parallel}.
The drawback of all these hand-engineered approaches is that they are not robust to real-world driving scenarios, thus requires tremendous engineering efforts to fine-tune it.

\textbf{Planning under uncertainty:} Planning methods for robust and safe driving in the presence of uncertainty have also been explored  \cite{bandyopadhyay2013intention,hardy2013contingency, zhan2016non}. Uncertainty in the intention of other actors is the main focus of \cite{bandyopadhyay2013intention,zhan2016non}. In \cite{hardy2013contingency}, possible future actions of other vehicles and collision probability are used to account for the uncertainty in obstacles positions. Compared to these approaches, our planner naturally handles uncertainty by learning a non-parametric cost function. %

\textbf{Holistic Models:} These models  provide interpretability. Chen \etal \cite{chen2015deepdriving} propose to learn a mapping from the sensor data to affordances, such as distance to left boundary/leading vehicle. This is then fed into a controller that generates steering command and acceleration. Sauer \etal \cite{sauer2018conditional} further propose a variant conditioned on direction command.
On the other hand, Luo \etal \cite{luo2018fast} propose a joint model for perception and prediction from raw LiDAR data and \cite{pmlr-v87-casas18a} extends it to predict each vehicle's intention.
All the methods above are trained for tasks that provide interpretable perception/prediction outputs to be used in motion planning. However, no feed-back is back-propagated from the motion planning module.

\begin{figure}[t]
\vspace{-0.3cm}
\begin{center}
  \includegraphics[height=4.5cm]{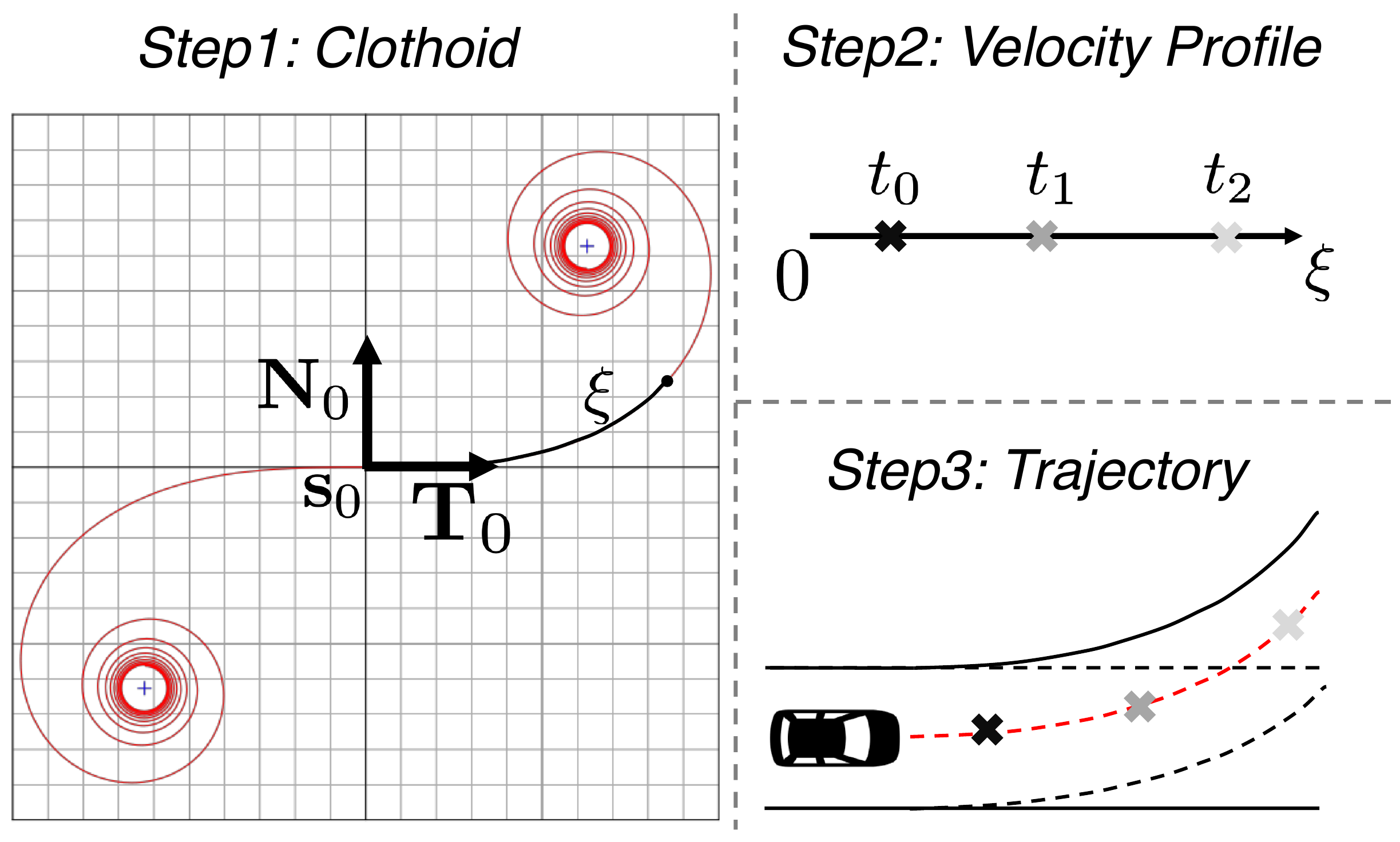}
\end{center}
\vspace{-0.3cm}
   \caption{Trajectory Representation. We first sample a set of parameters of a Clothoid to determine the shape of a trajectory. We then sample a velocity profile to determine how fast the SDV go along this trajectory. Combining these two, we can get a space-time trajectory.}
\label{fig:traj}
\vspace{-0.3cm}
\end{figure}

In this work, we take a holistic model approach and  take it one step further by designing a single neural network that takes raw sensors and dynamic map data as input and predicts the cost map for planning.
Compared with imitation learning approaches \cite{bojarski2016end,codevilla2018end, pomerleau1989alvinn} that directly regress a steer angle (from the raw data), our approach provides interpretability and handles  multimodality naturally. When compared with traditional planners which use manually designed cost functions built on top of perception and prediction systems, our model has the advantage of being jointly trained and thus learns representations that are optimal for the end-task. Furthermore, our model can handle uncertainty naturally (as this is represented in the cost) and does not require costly parameter tuning.

\section{Deep Structured  Interpretable Planner}

We propose an end-to-end learnable  motion planner that generates accurate space-time trajectories over a planning horizon of a few seconds.
Importantly, our model takes as input LiDAR point clouds and a high definition map  and produces  interpretable intermediate representations in the form of 3D detections and their future motion forecasted over the planning horizon. Our final output representation is  a space-time   cost volume that represents the ``goodness" of each possible location that the SDV can take within the planning horizon.
Our planner then scores a series of trajectory proposals using the learned cost volume and chooses the one with the minimum cost.

We train our model end-to-end with a multi-task objective. Our planning loss encourages the minimum cost plan to be similar to the trajectory performed by human demonstrators. Note that this loss is sparse as a ground-truth trajectory only occupies small portion of the space. As a consequence, learning with this loss alone is slow and difficult.  %
To mitigate this problem, we introduce an another perception loss that
 encourages the intermediate representations to produce accurate 3D detections and motion forecasting. This  ensures the interpretability of the intermediate representations and enables much faster learning. %

\subsection{Deep Structured Planning}

More formally, let $\mathbf{s} = \{\mathbf{s}^0, \mathbf{s}^1, \cdots, \mathbf{s}^{T-1}\}$ be a trajectory spanning over $T$ timesteps into the future, with $\mathbf{s}^t$  the location in bird's eye view (BEV) at the timestep $t$. We formulate the planning problem as a deep structured  minimization problem as follows
\begin{equation}
\mathbf{s}^* = \arg \min_\mathbf{s} \sum_t  c^t( \mathbf{s}^t)
\label{eq:obj}
\end{equation}
where $c^t$ is our learned cost volume indexed at the timestep $t$, which is a 2D tensor with the same size as our region of interest.
This minimization is approximated by sampling a set of physically valid trajectories \textbf{s}, and picking the one with minimum cost. Our model employs a convolutional network backbone to compute this cost volume. It first extracts features from both LiDAR and maps, and then feeds this feature map into two branches of convolution layers that  output 3D detection and motion forecasting as well as  the planning  cost volume respectively. In this section we describe our input representation and network in details.

\vspace{-0.3cm}
\paragraph{Input representation:}
\label{sub:input}
Our approach takes raw point clouds as inputs, captured by a LiDAR mounted on top of the SDV.
We employ $T'=10$ consecutive sweeps as observations, in order to infer the motion of all actors.
For those sweeps, we correct for ego-motion and bring the point clouds from the past 10 frames  to the same coordinate system centered at SDV's current location.
To make the input data amenable to standard convolutions, we follow  \cite{pmlr-v87-casas18a} and rasterize the space into a 3D occupancy  grid, where each voxel has a binary value indicating whether it contains a LiDAR point.
This results in  a 3D tensor of size $H$x$W$x$(ZT')$, where $Z, H, W$ represents the height and x-y spatial dimensions respectively. Note that we have concatenated timesteps along the $Z$ dimension, thus avoiding 3D convolutions which are memory and computation intensive. %

Access to a map is also a key for accurate motion planning, as we need to drive according to traffic rules (e.g., stop at a red light, follow the lane, change lanes only when allowed).
Towards this goal, we exploit HD maps that contain information about the semantics of the scene such as the location of lanes, their boundary  type (e.g., solid, dashed) and  the location of stop signs. %
Similar to \cite{pmlr-v87-casas18a}, we rasterize the map to form an $M$ channels tensor, where each channel represents a different map element, including road, intersections, lanes, lane boundaries, traffic lights, etc.
Our final input tensor is thus of size  $H$x$W$x$(ZT'+M)$.
\vspace{-0.3cm}
\paragraph{Backbone:}
\label{sub:net}
Our backbone is adapted from the detection network of  \cite{yang2018pixor} and  consists of five blocks. Each block has \{2, 2, 3, 6, 5\} \textit{Conv2D} layers with filter number \{32, 64, 128. 256, 256\}, filter size 3x3 and stride 1. There are \textit{MaxPool} layers after each of the first 3 blocks.
A multi-scale feature map is generated after the first 4 blocks as follows. %
We resize the feature maps from each of the first 4 blocks to 1/4 of the input size and concatenate them together similar to \cite{zhao2016pyramid}, in order to increase the  effective receptive field \cite{luo2016understanding}. These multi-scale features are then  fed into the $5$-{th} block.
The whole backbone has a downsampling rate of 4.

\vspace{-0.3cm}
\paragraph{Perception  Header:}
The perception header has two components formed of convolution layers, one for classification and one for regression. To reduce the variance of regression targets, we follow SSD \cite{liu2016ssd} and employ multiple predefined anchor boxes $a_{i,j}^k$ at each feature map location, where subscript $i, j$ denotes the location on the feature map and $k$  indexes over the anchors. In total, there are 12 anchors at each location, with different sizes, aspect ratios and orientations.
The classification branch  outputs a score $p_{i,j}^k$ for each anchor indicating the probability of a vehicle at each anchor's location. The regression branch also outputs regression targets for each anchor $a_{i,j}^k$ at different time-steps. %
This includes localization offset $l_x^t, l_y^t$, size $s_w^t, s_h^t$ and heading angle $a_{sin}^t, a_{cos}^t$. The superscript $t$ stands for time frame, ranging from $0$ (present) to $T-1$ into the future.
Regression is performed at every timesteps, thus producing motion forecasting for each vehicle.

\vspace{-0.3cm}
\paragraph{Cost Volume Head:}
The cost volume head consists of several convolution and deconvolution layers. To produce a cost volume $c$ at the same  resolution  as our bird-eye-view (BEV) input, we apply two deconvolution layers on the backbone's output with filter number \{128, 64\}, filter size 3x3 and stride 2. Each deconvolution layer is also followed by a convolution layer with filter number \{128, 64\}, filter size 3x3 and stride 1.
We then apply a final convolution layer with filter number $T$, which is our planning horizon. Each filter generates a cost volume $c^t$ for a future timestep $t$.  This allows us to evaluate the cost of any trajectory $\mathbf{s}$ by simply indexing in the cost volume $c$. In our experiments, we also clip the cost volume value between -1000 to +1000 after the network. Applying such bounds prevents the cost value shifting arbitrarily, and makes tuning hyper-parameters easier.
We next describe our output trajectory parameterization.

\subsection{Efficient Inference}
\label{sub:sampling}

Given the  input LiDAR sweeps and the HD map, we can compute the corresponding cost volume $c$ by feedforward convolutional operations as describe above.
The final trajectory can then be computed by minimizing Eq.~(\ref{eq:obj}). Note, however, that this optimization is NP hard\footnote{We expect the output trajectory of our planner is physically feasible. This introduces constraints on the solution set. Under these physical constraints, the optimization is NP hard.}. We thus rely on sampling to obtain a low cost trajectory.
Towards this goal, we sample a wide variety of trajectories that can be executed by the SDV and produce as final output the one with minimal cost according to our learned cost volume. %
In this section we describe how we efficiently sample physically possible trajectories during inference. Since the cost of a trajectory is computed by indexing from the cost volume, our planner is fast enough for real-time inference.

\vspace{-0.3cm}
\paragraph{Output Parameterization:}
\label{sub:traj}

A trajectory can be defined by the combination of  the spatial path (a curve in the 2D plane) and the velocity profile (how fast we go along this path). Sampling a trajectory as a set of points in $(x,y) \in \Re^2$ space is not a good idea, as a vehicle cannot execute all possible set of points in the cartesian space. This is due for example to the physical limits in speed, acceleration and turning angle. To consider these real-world constraints, we impose that the vehicle should follow a dynamical model.
In this paper, %
we employ the bicycle model \cite{paden2016survey}, which is widely used for planning in self-driving cars.  This model  implies that the curvature $\kappa$ of the vehicle's path is approximately proportional to the steering angle $\phi$ (angle between the front wheel and the vehicle):
$ \kappa = 2tan(\phi)/L \approx 2\phi/L, $
where $L$ is the distance between the front and rear axles of the SDV. This is a good approximation  as $\phi$ is usually small.

We then utilize a {\it Clothoid} curve,  also known as Euler spiral or  Cornu spiral, to represent the 2D path of the SDV \cite{shin1992path}. We refer the reader to  Fig.~\ref{fig:traj} for  an illustration. The curvature $\kappa$ of a point on this curve is proportional to its distance $\xi$ alone the curve from the reference point, i.e.,
$\kappa(\xi) = \pi \xi$.
Considering the bicycle model, this linear curvature characteristic corresponds to steering the front wheel angle with constant angular velocity.  %
The canonical form of a Clothoid can be defined as
\begin{equation}
  \mathbf{s}(\xi) = \mathbf{s_0} + a \left[ C\left(\frac{\xi}{a}\right)\mathbf{T_0} + S\left(\frac{\xi}{a}\right) \mathbf{N_0}\right]
  \label{eq:clothoid}
\end{equation}
\begin{equation}
  S(\xi) = \int_0^\xi sin \left(\frac{\pi u^2}{2}\right)du
\end{equation}
\begin{equation}
  C(\xi) = \int_0^\xi cos \left(\frac{\pi u^2}{2}\right)du
\end{equation}
Here, $\mathbf{s}(\xi)$ defines a Clothoid curve on a 2D plane, indexed by the distance $\xi$ to reference point $\mathbf{s_0}$, $a$ is a scaling factor,
$\mathbf{T_0}$ and $\mathbf{N_0}$ are the tangent and normal vector of this curve at point $\mathbf{s_0}$. $S(\xi)$ and $C(\xi)$ are called the \textit{Fresnel integral}, and can be efficiently computed.
In order to fully define a trajectory, we also need a longitudinal velocity $\dot{\xi}$ (velocity profile) that specifies the SDV motion along the path $\mathbf{s(\xi)}$: $\dot{\xi}(t) = \ddot{\xi}t+\dot{\xi}_0,$
where $\dot{\xi}_0$ is the initial velocity of the SDV and $\ddot{\xi}$ is a constant forward acceleration. Combining this and (\ref{eq:clothoid}), we can obtain the trajectory points $\mathbf{s}$ in Eq.~(\ref{eq:obj}).

\vspace{-0.3cm}
\paragraph{Sampling:}
Since we utilize Clothoid curves, sampling a path corresponds to sampling the scaling factor $a$ in Eq.~(\ref{eq:clothoid}).
Considering the city driving speed limit of 15m/s,  we sample $a$ uniformly from the range of 6 to 80m. Once $a$ is sampled, the shape of the curve is fixed.\footnote{We also sample a binary random variable indicating it's a canonical Clothoid or a vertically flipped mirror. They correspond with turning left or right respectively. } We then  use the initial SDV's  steering angle (curvature) to find the corresponding position on the curve.
Note that Clothoid curves cannot  handle circle and straight line trajectories well, thus we sample them separately. The probability of using  straight-line, circle and Clothoid curves are 0.5, 0.25, 0.25 respectively. Also, we only use a single Clothoid segment to specify the path of SDV which we think is enough for the short planning horizon.
In addition, we sample constant accelerations $\ddot{\xi}$ ranging from $-5m/s^2$ to $5m/s^2$ which specifies the SDV's velocity profile.
Combining sampled curves and velocity profiles, we can project the trajectories to discrete timesteps and obtain the corresponding waypoints (See Fig~\ref{fig:traj}) for which to evaluate the learned cost.

\begin{table*}[t]
\centering
\small
\begin{tabular}{ccccccccccccc}
    \specialrule{.2em}{.1em}{.1em}
    Method & \multicolumn{3}{c}{L2 (m)} & \multicolumn{6}{c}{Collision Rate (\%)} & \multicolumn{3}{c}{Lane Violation (\%)}\\
    & 1.0s & 2.0s & 3.0s & 0.5s & 1.0s & 1.5s & 2.0s & 2.5s & 3.0s & 1.0s & 2.0s & 3.0s \\
    \hline
    Ego-motin & 0.281 & 0.900 & 2.025 & \textbf{0.00} & \textbf{0.01} & 0.20 & 0.54 & 1.04 & 1.81 & 0.51 & 2.72 & 6.73 \\
    IL & \textbf{0.231} & \textbf{0.839} & \textbf{1.923} & \textbf{0.00} & \textbf{0.01} & 0.19 & 0.55 & 1.04 & 1.72 & 0.44 &  2.63 & 5.38\\
    Acc & 0.403	& 1.335 & 2.797 & 0.05 & 0.12 & 0.27 & 0.53 & 1.18 & 2.39 & \textbf{0.24} & \textbf{0.46} & \textbf{0.64} \\
    Manual Cost & 0.402 & 1.432 & 2.990 & \textbf{0.00} & 0.02 & 0.09 & 0.22 & 0.79 & 2.21 & 0.39 & 2.73 & 5.02 \\
    Ours-NMP & 0.314 & 1.087 & 2.353 & \textbf{0.00} & \textbf{0.01} & \textbf{0.04} & \textbf{0.09} & \textbf{0.33} & \textbf{0.78} & 0.35& 0.77 & 2.99 \\
    \specialrule{.1em}{.05em}{.05em}
\end{tabular}
\caption{Planning Metrics}
\label{table:planning}
\end{table*}

\subsection{End-to-End Learning}
\label{sec:learning}

Our ultimate goal is to plan a safe %
 trajectory while following the rules of traffic.
We want the model to understand where obstacles are and  where they will be in the future in order  to avoid collisions. Therefore, we use a multi-task training with supervision from detection,  motion forecasting as well as human driven trajectories for the ego-car.
Note that we do not have supervision for cost volume.  We thus adopt max-margin loss to push the network to learn to discriminate between good and bad trajectories. %
The overall loss function is then:
\begin{equation}
  \mathcal{L} = \mathcal{L}_{\text{perception}} + \beta \mathcal{L}_{\text{planning}}.
\end{equation}
This multi-task loss not only directs the network to extract useful features, but also make the network output interpretable results. This is crucial for self-driving as it helps understand failure cases and improves the system. %
In the following, we describe  each  loss in more details.

\vspace{-0.3cm}
\paragraph{Perception Loss:} Our perception loss includes classification loss, for distinguishing a vehicle from the background, and regression loss, for generating precise object bounding boxes. For each predefined anchor box, the network outputs a classification score as well as several regression targets. This classification score $p_{i,j}^k$ indicates the probability of existence of   a vehicle at this anchor. We employ a cross-entropy loss for the classification defined as %
\begin{equation}
  \mathcal{L}_{cla} = \sum_{i, j, k}\left(  q_{i,j}^k \log p_{i,j}^k + (1-q_{i,j}^k)\log (1-p_{i,j}^k)\right),
\end{equation}
where $q_{i,j}^k$ is the class label for this anchor (i.e., $q_{i,j}^k=1$ for vehicle and $0$ for background). The regression outputs include information of position, shape and heading angle at each time frame $t$, namely
$$ l_x = \frac{x^{a}-x^{l}}{w^{a}} \quad l_y = \frac{y^{a} - y^{l}}{h^{a}},$$
$$ s_w = \log \frac{w^{l}}{w^{a}} \quad s_h = \log \frac{h^{l}}{h^{a}},$$
$$ a_{sin} = sin(\theta^{a} - \theta^{l}) \quad a_{cos} = cos(\theta^{a} - \theta^{l}),$$
where superscript $a$ means anchor and $l$ means label.
We use a weighted smooth L1 loss over all these outputs.
The overall \textit{perception loss} is
\begin{equation}
  \mathcal{L}_{perception} = \sum \left( \mathcal{L}_{cla} + \alpha \sum_{t=0}^{T} \mathcal{L}^t_{reg} \right).
\end{equation}
Note that the regression loss is summed over all vehicle correlated anchors, from the current time frame to our prediction horizon $T$. Thus it teaches the model to predict the position of vehicles at every time frame.

To find the training label for each anchor, we associate it to its neighboring ground-truth bounding box, similar to \cite{liu2016ssd, luo2018fast}. In particular, for each anchor, we find all the ground-truth boxes with intersection over union (IoU) higher than $0.4$. We associate the highest one among them to this anchor, and compute the class label and regression targets accordingly. We also associate any non-assigned ground-truth boxes with their nearest neighbor. The remaining anchors are treated as background, and are not  considered in the regression loss. Note that one ground-truth box may associate to multiple anchors, but one anchor can at most be associated with one ground-truth box. During training, we also apply hard negative mining to overcome imbalance between positive and negative samples.

\vspace{-0.2cm}
\paragraph{Planning Loss:}
Learning a reasonable cost volume is challenging as we do not have ground-truth. To overcome this difficulty, we minimize the max-margin loss where we use the ground-truth trajectory as a positive example, and randomly sampled trajectories as negative examples.
The intuition behind is to encourage the ground-truth trajectory to have the minimal cost, and others to have higher costs. More specifically, assume we have a ground-truth trajectory $\{(x^t, y^t)\}$ for the next $T$ time steps, where $(x^t, y^t)$ is the position of our vehicle at the $t$ time step. Define the cost volume value at this point $(x^t, y^t)$ as  $\hat{c}^t$. Then, we sample $N$ negative trajectories, the $i^{th}$ among which is $\{(x_i^t, y_i^t)\}$ and the cost volume value at these points are $c_i^t$. The sampling procedure for negative trajectories is similar as we described in Section. \ref{sub:sampling}, except there is 0.8 probability that the negative sample doesn't obey SDV's initial states, e.g. we randomly sample a velocity to replace SDV's initial velocity. This will provide easier negative examples for the model to start with. The overall max-margin loss is defined as
\vspace{-0.2cm}
\begin{equation}
  \label{eq:costmap_obj}
  \mathcal{L}_{\text{planning}} = \sum_{\tiny{\{(x^t, y^t)\} }}\left( \max_{1 \leq i \leq N} \left( \sum_{t=1}^T \left[  \hat{c}^t - c_i^t + d_i^t +  \gamma_i^t \right]_{+} \right) \right)
\end{equation}
The inner-most summation denotes the discrepancy between the ground-truth trajectory and one negative trajectory sample, which is a sum of per-timestep loss.
$[]_+$ represents a ReLU function. This is designed to be inside the summation rather than outside, as it can prevent the cost volume at one time-step from dominating the whole loss.
$d_i^t$ is the distance between negative trajectory and ground-truth trajectory $|| (x^t, y^t) - (x_i^t, y_i^t)||_2$, which is used to encourage negative trajectories far from the ground-truth trajectory to have much higher cost.
$\gamma_i^t$ is the traffic rule violation cost, which is a constant  if and only if the negative trajectory $t$ violates traffic rules at time $t$, e.g. moving before red-lights, colliding with other vehicles etc. This is used to determined how `bad' the negative samples are, as a result, it will penalize those rule violated trajectories more severely and thus avoid dangerous behaviors.
After computing the discrepancy between the ground-truth trajectory and each negative  sample, we only optimize the worst case by the $\max$ operation. This encourages the model to learn a cost volume that discriminates good trajectories from bad ones.

\begin{table*}[t]
\vspace{-0.3cm}
\centering
\small
\begin{tabular}{ccccccccccccccccc}
  \specialrule{.2em}{.1em}{.1em}

  Method & \multicolumn{4}{c}{L2 along trajectory (m)} & \multicolumn{4}{c}{L2 across trajectory (m)} & \multicolumn{4}{c}{L1 (m)} & \multicolumn{4}{c}{L2 (m)}\\

  & 0s & 1s & 2s & 3s & 0s & 1s & 2s & 3s & 0s & 1s & 2s & 3s & 0s & 1s & 2s & 3s\\\hline

  FaF\cite{luo2018fast} &  0.29 & 0.49 & 0.87 & 1.52 & 0.16 & 0.23 & 0.39  & 0.58 & 0.45 & 0.72 & 1.31 & 2.14 & 0.37 & 0.60 & 1.11 & 1.82\\
  IntentNet\cite{pmlr-v87-casas18a} & 0.23 & 0.42 & 0.79 & 1.27 & 0.16 & 0.21 & 0.32 & 0.48 & 0.39 & 0.61 & 1.09 & 1.79 & 0.32 & 0.51 & 0.93 & 1.52\\

  Ours-NMP & \textbf{0.21} & \textbf{0.37} & \textbf{0.69} & \textbf{1.15} & \textbf{0.12} & \textbf{0.16} & \textbf{0.25} & \textbf{0.37} & \textbf{0.34} & \textbf{0.54} & \textbf{0.94} & \textbf{1.52} & \textbf{0.28} & \textbf{0.45} & \textbf{0.80} & \textbf{1.31}\\
  \specialrule{.1em}{.05em}{.05em}

\end{tabular}
\caption{Motion Forecasting Metric}
\label{table:prediction_result}
\end{table*}

\section{Experiments}

\begin{table*}[t]
\centering
\small
\begin{tabular}{ccccccccccccccccc}
  \specialrule{.2em}{.1em}{.1em}
  ID &
  \multicolumn{2}{c}{Loss} & \multicolumn{2}{c}{Input} & %
  Penalty & \multicolumn{2}{c}{mAP@IoU} & \multicolumn{3}{c}{Prediction L2 (m)} & \multicolumn{3}{c}{Collision Rate (\%)}& \multicolumn{3}{c}{Traffic Violation (\%)}\\
  & Det & Plan & 5 & 10 & & 0.5 & 0.7 & 1s & 2s & 3s & 1s & 2s & 3s & 1s & 2s & 3s \\
  \hline
  1 &\checkmark & & & \checkmark & & 94.1 & \textbf{81.3} & 0.48 & 0.84 & 1.34 & -& -& -& -& -& - \\
  2 & & \checkmark & & \checkmark & & - & - & -& -& -& \textbf{0.01} & 0.23 & 1.42 &  0.37 & 1.06 & 3.85 \\
  3 & \checkmark & \checkmark & \checkmark & & \checkmark & 93.6 & 80.1 & 0.46 & 0.83 & 1.35 & \textbf{0.01} & 0.15 & 0.93 & 0.36 & 0.86 & 3.09 \\
  4 & \checkmark & \checkmark & & \checkmark &  & \textbf{94.2} & 81.1 & \textbf{0.45} & \textbf{0.80} & \textbf{1.30} & \textbf{0.01} & 0.29 & 1.40 & 0.36 & 1.02 & 3.26\\
  5 & \checkmark & \checkmark & & \checkmark &  \checkmark & \textbf{94.2} & 81.1 & \textbf{0.45} & \textbf{0.80} & 1.31 &\textbf{0.01} &\textbf{0.09} &\textbf{0.78} & \textbf{0.35}& \textbf{0.77} & \textbf{2.99} \\
  \specialrule{.1em}{.05em}{.05em}

\end{tabular}
\caption{Ablation Study. We compare effects of different supervisions, different input horizons and different training losses. ID denotes model id which we use for clarity and brevity.}
\label{table:ablation_result}
\vspace{-0.2cm}
\end{table*}

In this section, we evaluate our approach on a large scale real-world driving dataset. The dataset was collected over multiple cities across North America. It consists of 6,500 scenarios with about 1.4 million frames, the training set consists of 5,000 scenarios, while validation and test have 500 and 1,000 scenarios respectively.
Our dataset has annotated 3D bounding boxes of vehicles for every 100ms.  %
For all experiments, we utilize the same spatial region, which is centered at the SDV, with 70.4 meters both in front and back, 40 meters to the left and right, and height from -2 meters to 3.4 meters. This corresponds to a 704x400x27 tensor. Our input sequence is 10 frames at 10Hz, while the output is 7 frames at 2Hz, thus resulting in  a planning horizon of 3 seconds.

In the following, we first show quantitative analysis on planning on a wide variety of metrics measuring collision, similarity to human trajectory and traffic rule violation. Next we demonstrate the interpretability of our approach, through quantitative analysis of detection and motion forecasting, as well as visualization of the  learned cost volume. Last, we provide an ablation study to show the effects of different loss functions and  different temporal history lengths.

\subsection{Planning Results}

We evaluate a wide variety of planning metrics.
\textbf{L2 Distance to Real Trajectory}: This evaluates how far away the planned trajectory is from the real executed trajectory. Note that the real trajectory is just one of the many possible trajectories that a human could do, and thus this metric is not perfect. \textbf{Future Potential Collision Rate}: This is used to see if the planned trajectory will overlap with other vehicles in the future. For a given timestep t, we compute the percentage of occurrence of collisions up to time t, thus lower number is preferred. \textbf{Lane Violation}: this metric counts the percentage  of planned trajectories \textit{crossing} a solid yellow line. Note that lower is better, and here \textit{crossing} is defined  if the SDV touches the line.

\begin{figure*}[tb]
\vspace{-0.4cm}
\centering
\setlength{\tabcolsep}{1pt}
\begin{tabular}{ccc}
&\includegraphics[width=0.32\linewidth]{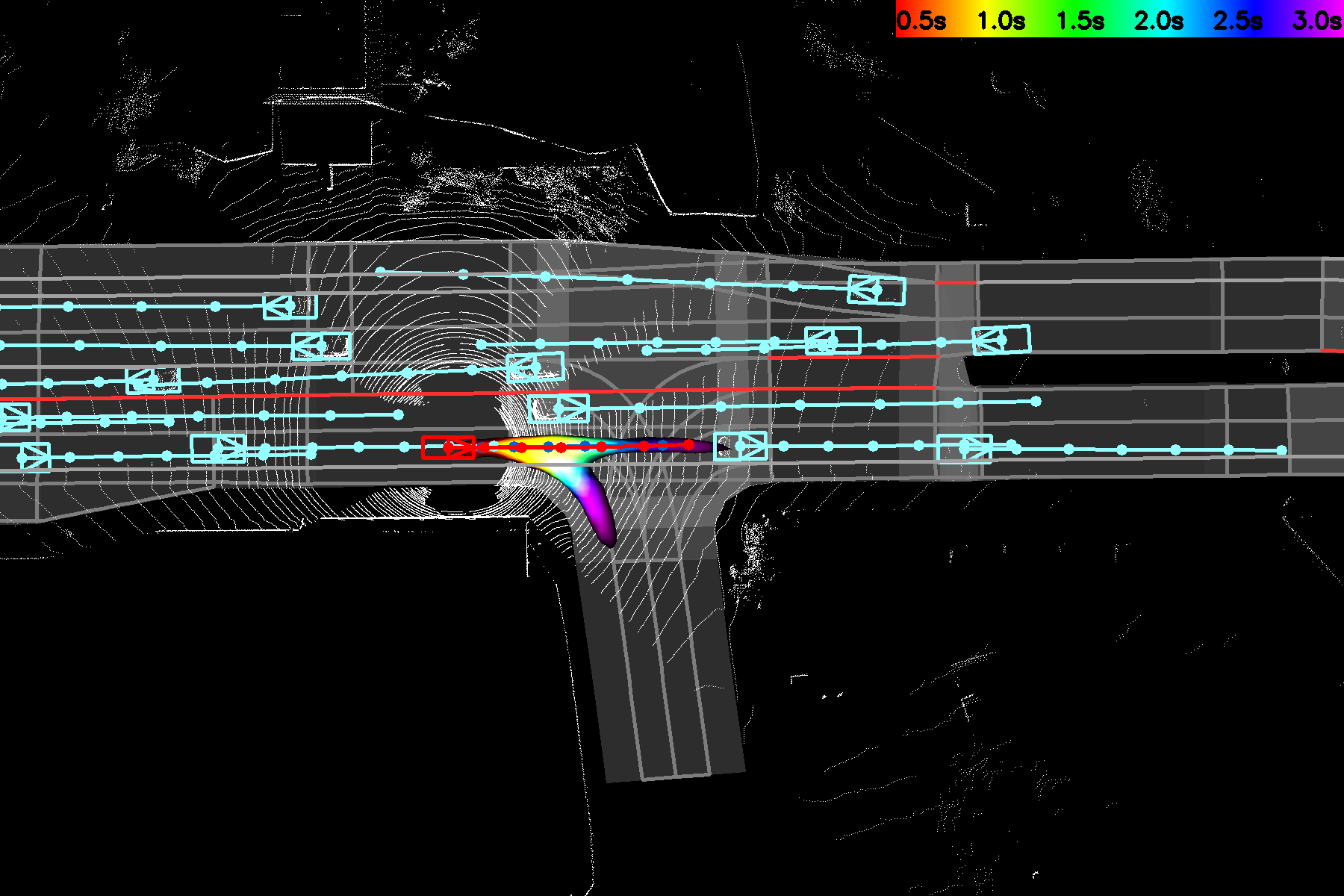}
\includegraphics[width=0.32\linewidth]{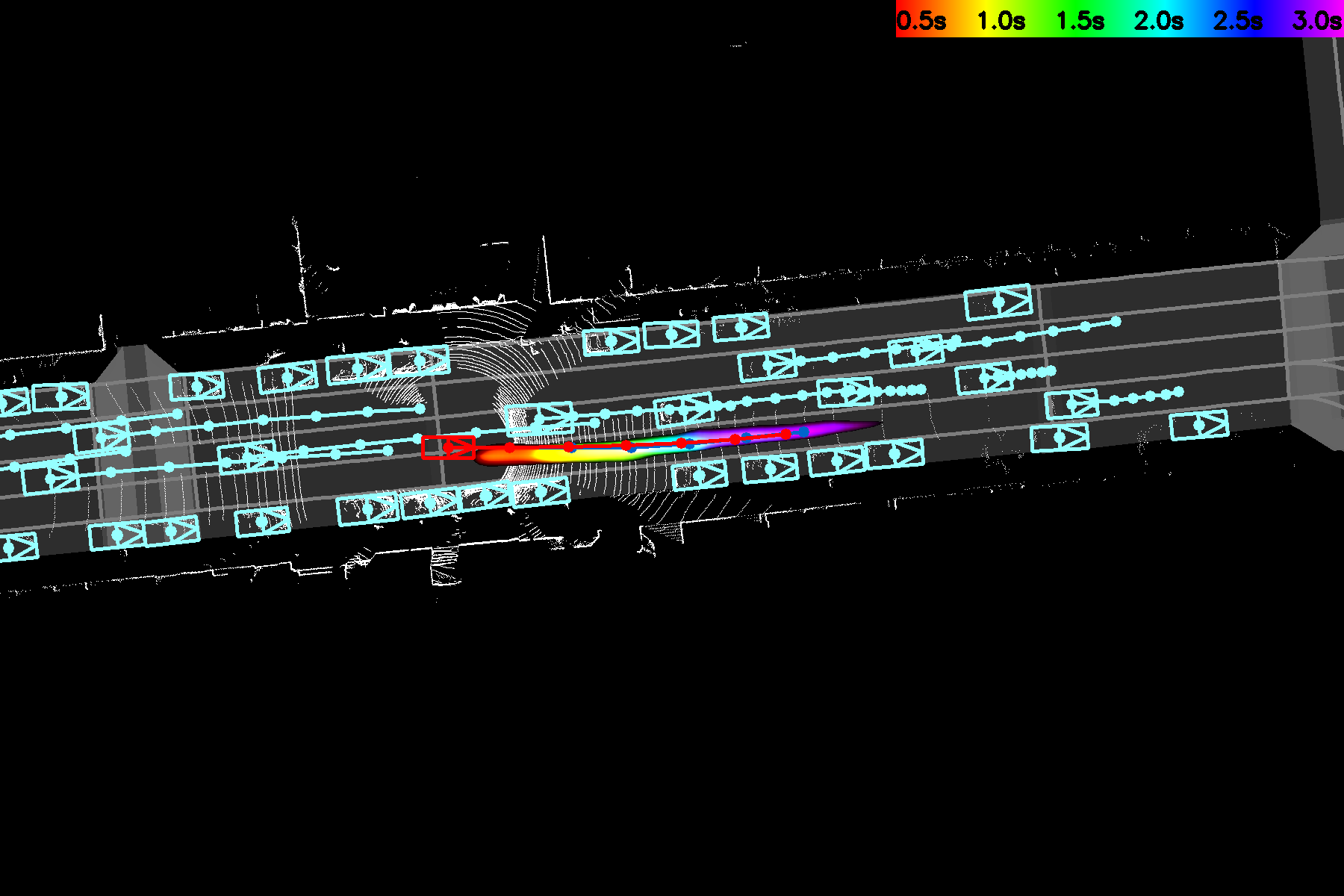}
\includegraphics[width=0.32\linewidth]{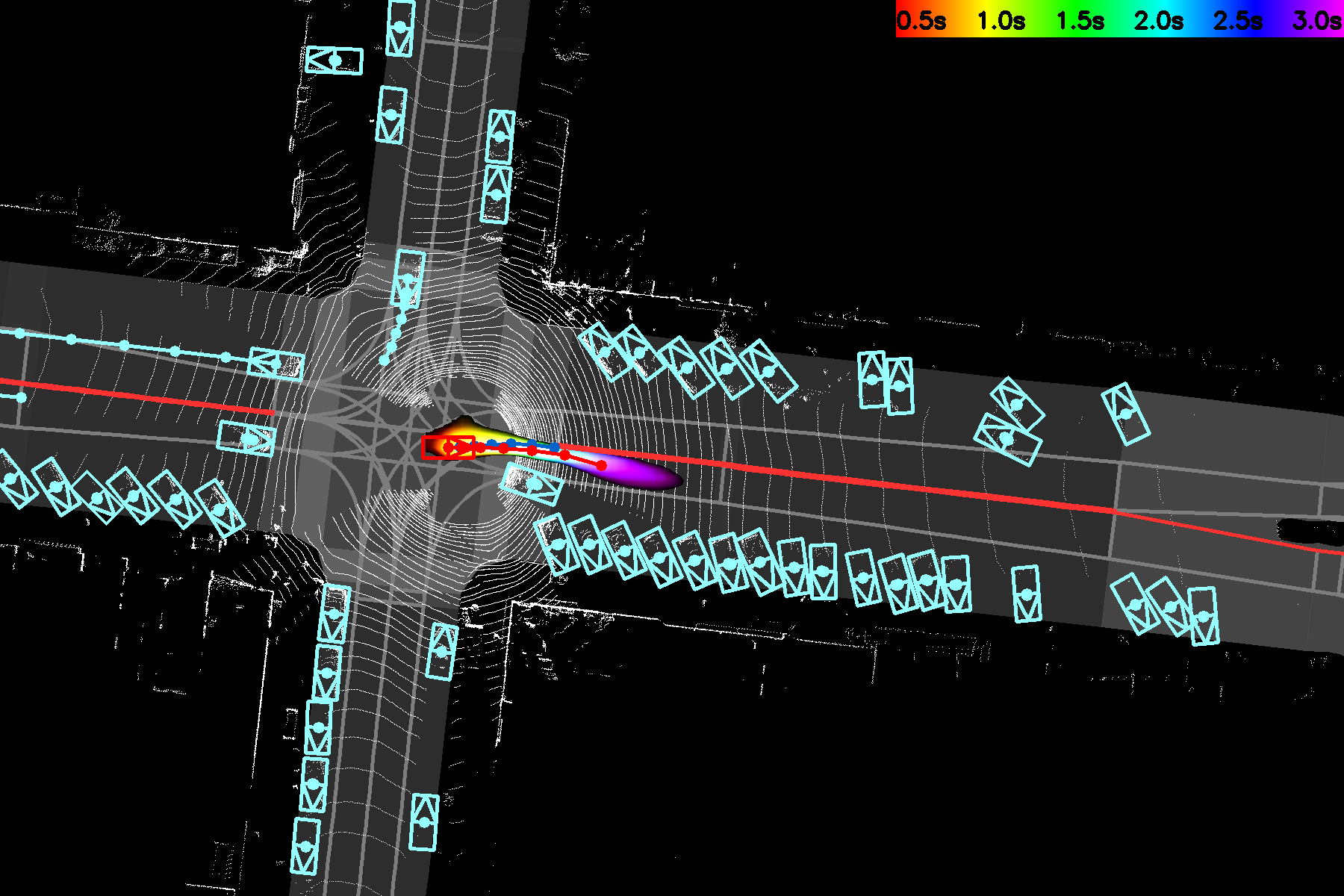}\\

&\includegraphics[width=0.32\linewidth]{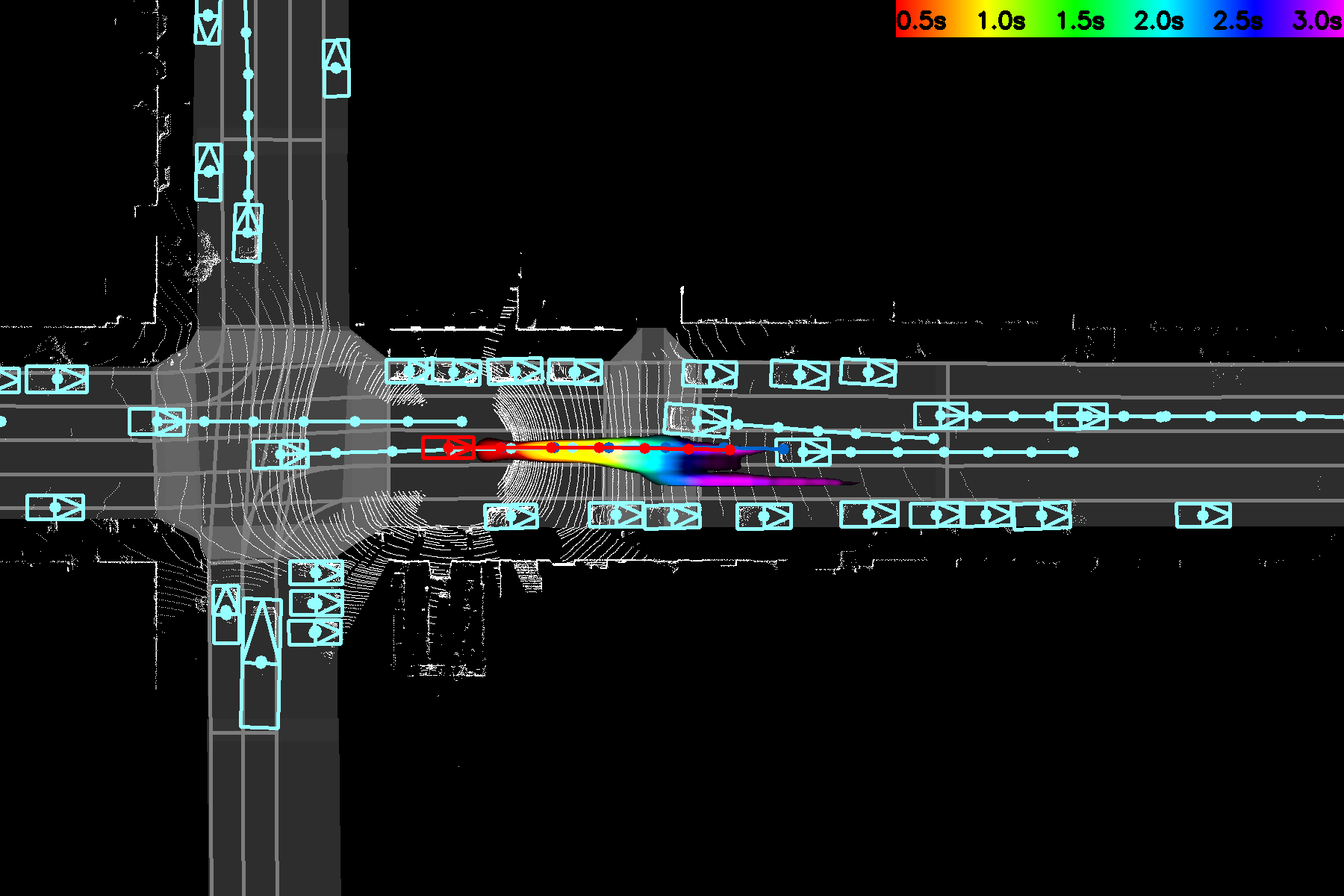}
\includegraphics[width=0.32\linewidth]{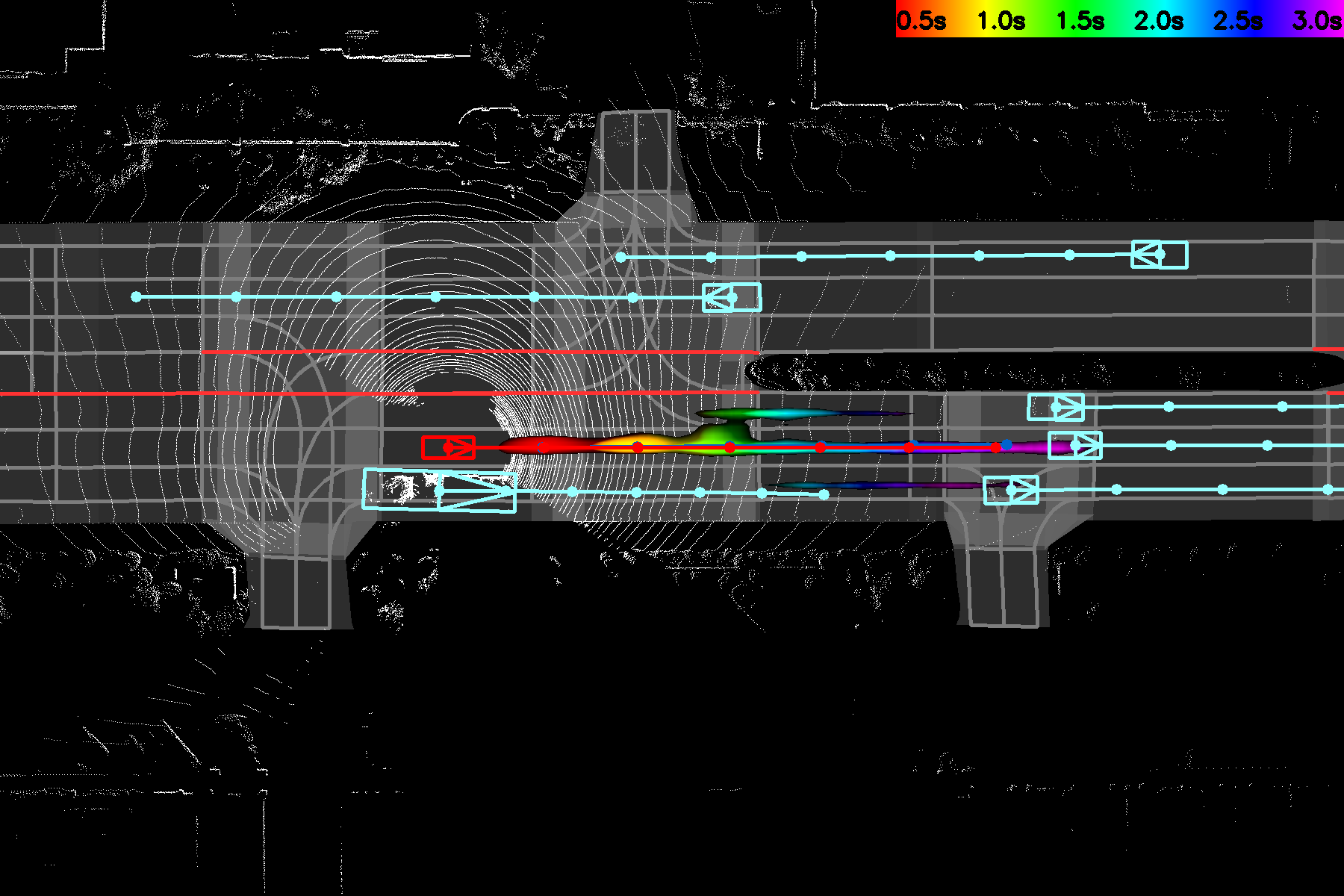}
\includegraphics[width=0.32\linewidth]{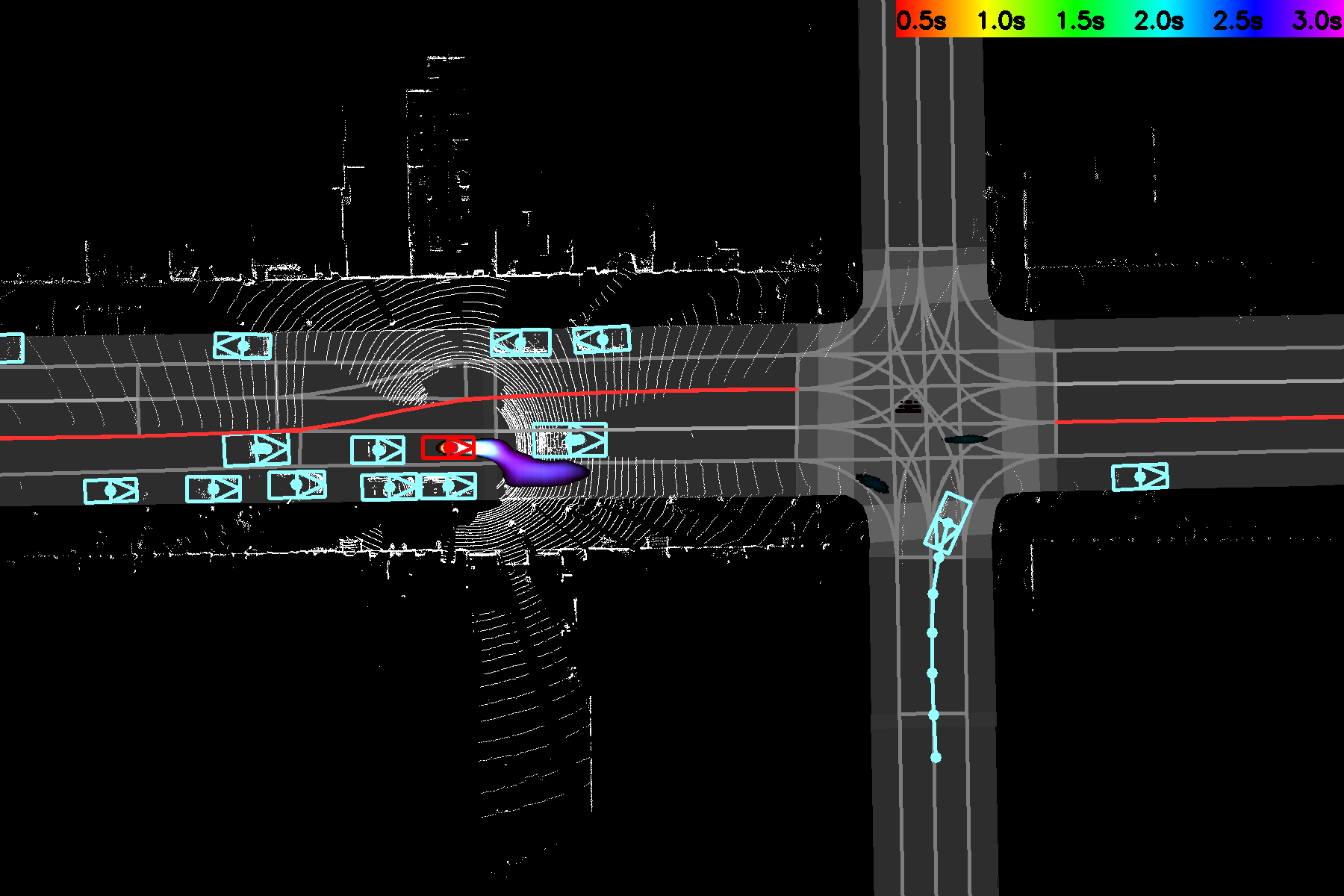}\\

\end{tabular}
\caption{\textbf{Cost Volume across Time} We shown planned trajectory in red and ground-truth in blue. We overlay lower cost region for different timesteps in the same figure, using different colors (indicated by legend). Detection and corresponding prediction results are in cyan. (best view in color)}
\vspace{-0.2cm}
\label{fig:vis}
\end{figure*}

We implement many  baselines for comparison including:
\textbf{Ego-motion forecasting (Ego-motion)}: Ego-motion provides a strong cue of how the SDV would move in the future.  This baselines takes only SDV's past position as input and uses a 4-layer MLP to predict the future locations.
\textbf{Imitation Learning (IL)}: We follow the imitation learning framework \cite{bojarski2016end,codevilla2018end, pomerleau1989alvinn}, and utilize  a deep network to extract features from raw LiDAR data and rasterized map. For fair comparison, we use the same backbone described (Sec.~\ref{sub:net}) and same input parameterization (Sec.~\ref{sub:input}) than our approach. In addition, the same MLP from \textit{Ego-motion forecasting} baseline is used to extract features from ego-motion. These two features are then concatenated and fed into a 3 layer MLP to compute the final prediction.
\textbf{Adaptive Cruise Control (ACC)}: This baseline implements the simple behavior of following the leading vehicle. The vehicle follows the lane center-line, while adaptively adjusting its speed to maintain a safe distance from the vehicle ahead. When there is no lead vehicle, a safe speed limit is followed. Traffic controls (traffic lights, stop signs) are observed as a stationary obstacle, similar to a stopped lead vehicle.
\textbf{Plan w/ Manual Cost (Manual)}: This baselines uses the same trajectory parameterization and sampling procedure as our approach. However it utilizes a manually designed cost using perception and motion forecasting outputs. In detail, we rasterize all possible roads the SDV can take going forward and set it to a low cost of 0; all  detected objects's bounding box defines area of a high cost set to 255; cost of any other area is set to a default value 100.
This baseline is designed to show the effectiveness of our learned cost volume as it utilize the same sampling procedure as our approach but just a different cost volume.

As shown in  Tab.~\ref{table:planning}, our approach has lower future collision rate at all timesteps by a large margin. Note that Ego-motion and IL baselines give lower L2 numbers as they optimize directly for this metric, however they are not good from planning perspective as they have difficulty reasoning about other actors and collide frequently with them. Comparing to the manual cost baseline and ACC, we achieve both better regression numbers and better collision rates, showing the advantage of our learned cost volume over manual a designed cost. For lane violation,  ACC is designed to follow the lane, thus it has about 0 violation by definition. Comparing to other baselines, we achieve much smaller violation number, showing our model is able to reason and learn from the map.
\subsection{Interpretability}
Interpretability is crucial for self-driving as it can help understand failure cases. We showcase the interpretability of our approach by showing  quantitative results on 3D detection and motion forecasting and visualization our learned cost-map for all timesteps into the future.

\begin{table}[t]
\centering
\small
\begin{tabular}{cccccc}
  \specialrule{.2em}{.1em}{.1em}

  Method & \multicolumn{5}{c}{Detection mAP @ IoU (pts $\geq$ 1)}\\
  & 0.5 & 0.6 & 0.7 & 0.8 & 0.9 \\
  \hline
  MobileNet\cite{howard2017mobilenets} & 86.1 & 78.3 & 60.4 & 27.5 & 1.1\\
  FaF\cite{luo2018fast} & 89.8 & 82.5 & 68.1 & 35.8 & 2.5\\
  IntentNet\cite{pmlr-v87-casas18a} & \textbf{94.4} & 89.4 & 75.4 & 43.5 & 3.9\\
  Pixor\cite{yang2018pixor} & 93.4 & 89.4 & 78.8 & 52.2 & \textbf{7.6}\\

  Ours-NMP & 94.2 & \textbf{90.8} & \textbf{81.1} & \textbf{53.7} & 7.1\\
  \specialrule{.1em}{.05em}{.05em}

\end{tabular}
\caption{Detection mAP Result}
\label{table:detection_result}
\end{table}

\vspace{-0.4cm}
\paragraph{Detection:}
We compare  against several state-of-the-art real-time detectors, validating that our holistic model understand the environment. Our baselines include a MobileNet adapted from \cite{howard2017mobilenets}, FaF\cite{luo2018fast},
IntentNet\cite{pmlr-v87-casas18a} and Pixor\cite{yang2018pixor}, which are specifically designed for LiDAR-based  3D object detection.
The metric is mAP with different IoU thresholds, and vehicles without LiDAR points are not considered. As shown in  Tab.~\ref{table:detection_result},  our model archives best results on 0.7 IoU threshold, which is the metric of choice  for self-driving. Qualitative results can also be found in Fig.~\ref{fig:vis}.

\vspace{-0.3cm}
\paragraph{Motion Forecasting:}
Tab.~\ref{table:prediction_result} shows quantitative  motion forecasting results, including L1 and L2 distance to ground-truth locations. We also provides the L2 distance from our predictions to the ground-truth position along and perpendicular to the ground-truth trajectory. These help explain if the error is due to wrong velocity or direction estimation. We use baselines from \cite{pmlr-v87-casas18a, luo2018fast}, which are  designed for motion forecasting with raw LiDAR data.
Our model performs better in all metric and all time steps. Note that  IntentNet uses high-level intentions as additional information for training.  Qualitative results are shown in Fig.\ref{fig:vis}.

\vspace{-0.3cm}

\paragraph{Cost Map Visualization:} In Fig.~\ref{fig:vis}, we visualize a few different driving scenarios. Each figure gives a top-down view of the scene, showing the map, LiDAR point clouds, detection, motion forecasting and planning results including learned cost map.
Each figure represents one example, where we overlay the cost map from different timesteps. We use different color to represent the lower cost region for different timesteps (indicated by color legend).
As we can see, our model learns to produce a time-dependent cost map. In particular, the first column demonstrates multi-modality, second column shows lane-following in heavy traffic and the last column shows collision avoidance.

\subsection{Ablation Study}

We conduct ablation studies and report the results in Table \ref{table:ablation_result}. Our best model is Model 5, comparing to Model 1 which is optimized only for detection and motion forecasting, it achieves similar performance
in terms of detection and motion forecasting. Model 2 trains directly with planning loss only, without the supervision of object bounding boxes and performs worse. Model 3 exploits different input length, where longer input sequence gives better results.
Model 4 is trained without the traffic rule penalty $\gamma$ in Eq.~\ref{eq:costmap_obj}. It performs worse on planning, as it has no prior knowledge to avoid collision.

\section{Conclusion}

We have proposed a neural motion planner that learns to drive safely while following traffic rules.
We have designed a holistic model that takes  LiDAR data and an HD map and produces  interpretable intermediate representations in the form of 3D detections and their future trajectories, as well as a cost map defining the goodness of each position that the self-driving car  can take within the planning horizon. Our planer  then sample a set of  physically possible trajectories and chooses the one with the minimum learned cost.
We have demonstrated the effectiveness of our approach in very complex real-world  scenarios  in several cities of North America and show how we can learn to drive accurately.

{\small
\bibliographystyle{ieee_fullname}
\bibliography{ourbib_clean}
}

\end{document}